%% file: ms.tex
\documentclass{article}

\usepackage{PRIMEarxiv}

\usepackage[utf8]{inputenc} 
\usepackage[T1]{fontenc}    
\usepackage{hyperref}       
\usepackage{url}            
\usepackage{booktabs}       
\usepackage{amsfonts}       
\usepackage{nicefrac}       
\usepackage{microtype}      
\usepackage{lipsum}
\usepackage{fancyhdr}       
\usepackage{graphicx}       
\graphicspath{{media/}}     

\usepackage[round,authoryear,elide]{natbib}
\usepackage{tabularx} 
\usepackage{adjustbox}
\usepackage{pdflscape}
\usepackage{verbatim}
\usepackage{tablefootnote}
\usepackage{hyperref}
\usepackage[hyphenbreaks]{breakurl}
\usepackage{xurl}
\usepackage{amsmath,amsfonts,amssymb}

\usepackage{array}
\usepackage{multirow}

\usepackage{graphicx}

\usepackage[normalem]{ulem}
\newcommand{\todo}[1]{\textbf{\textcolor{red}{#1}}}

\newcommand{\OC}[1]{\textcolor{magenta}{#1}}

\newcommand{\MW}[1]{\textcolor{orange}{#1}}

\usepackage{afterpage}

\makeatletter
\newcommand*{\changepagecolor}{%
  \@ifnextchar[\@changepagecolor@i\@changepagecolor@ii
}
\def\@changepagecolor@i[#1]#2{%
  \@changepagecolor@do{[{#1}]{#2}}%
}
\newcommand*{\@changepagecolor@ii}[1]{%
  \@changepagecolor@do{{#1}}%
}
\newcommand*{\@changepagecolor@do}[1]{%
  \begingroup
    \offinterlineskip
    \hbox to 0pt{%
      \kern-\paperwidth
      \vtop to 0pt{%
        \color#1%
        \hrule width 2\paperwidth height \paperheight
        \vss
      }%
      \hss
    }%
  \endgroup
  \afterpage{\pagecolor#1}%
}
\makeatother

\usepackage{tikz}
\usetikzlibrary{shapes.geometric,decorations.pathreplacing,arrows,positioning}
\usepackage[utf8]{inputenc}
\usepackage[T1]{fontenc}
\usepackage{lmodern}

\tikzset{basic/.style={text width=1em,text badly centered}}
\tikzset{node/.style={basic,circle}}

\tikzset{ba/.style={draw=black,arrows=-latex, thick}} 

\newcounter{example}[section]

\newcounter{eq}[section]
\newenvironment{eq}[1][]{\refstepcounter{eq}\par\medskip
  \noindent \textbf{Equation~\theeq. #1} \rmfamily}{\medskip}

\title{Bias Mitigation Methods for Binary Classification Decision-Making Systems: Survey and Recommendations
}

\author{
  Madeleine Waller, Odinaldo Rodrigues, Oana Cocarascu \\
  Department of Informatics, King's College London, UK \\
  \texttt{\{madeleine.waller, odinaldo.rodrigues, oana.cocarascu\}@kcl.ac.uk} \\
}

\begin{document}
\maketitle

\begin{abstract}
Bias mitigation methods for binary classification decision-making systems have been widely researched due to the ever-growing importance of designing fair machine learning processes that are impartial and do not discriminate against individuals or groups based on protected personal characteristics. In this paper, we present a structured overview of the research landscape for bias mitigation methods, report on their benefits and limitations, and provide recommendations for the development of future bias mitigation methods for binary classification.

\end{abstract}


\include{macros}

\section{Introduction\label{Introduction}}

Artificial Intelligence (AI), and in particular machine learning (ML) techniques, have been used in decision-making systems which typically rely on historical data for training. However, the data may contain biases against groups or individuals with certain characteristics, which can lead to discriminatory or unfair decisions. Such decision-making systems are used in both private and public sectors. Organisations are increasingly relying on these systems to reduce workload and free up resources   \citep[e.g.][]{deloitteAIAutomatedDecision2021,enginAlgorithmicGovernmentAutomating2019}, and local governments, in particular social services,\footnote{\url{https://www.local.gov.uk/using-predictive-analytics-local-public-services}} have deployed systems to predict a score of how at risk a child is of neglect or abuse \citep{churchSearchSilverBullet2017}. The potential harmful impact of these systems is therefore immense.

There have been several recent examples of unfair decisions made by machine learning models in different domains, such as criminal justice~\citep{thepartnershiponaiReportAlgorithmicRisk2019}, recruitment,\footnote{\url{https://www.reuters.com/article/us-amazon-com-jobs-automation-insight-idUSKCN1MK08G}} and social services~\citep{gillinghamDecisionSupportSystems2019}. Algorithms used to predict if criminals will re-offend are used across many US states and yet an analysis of a popular tool, COMPAS \citep{northpointePractitionerGuideCOMPAS2019}, showed that black defendants were identified incorrectly as re-offending at a higher rate than white defendants \citep{larsonHowWeAnalyzed2016}. A recruitment tool used by Amazon was shown to be biased against women \citep{jeffreydastinAmazonScrapsSecret2018}, reinforcing historical biases due to the small sample size of women who had previously been hired.

 Ensuring AI systems are fair to individuals and communities is an important cross-disciplinary issue which must consider the context and application of the systems deployed \citep{wallerWhyPredictiveAlgorithms2020}. A system is considered to be fair if it does not discriminate based on protected personal characteristics such as race, sex, religion, etc. Discrimination may arise from the dataset used to train decision-making systems; specifically, unwanted bias occurs when the system favours or gives advantage to historically favoured groups or its outputs directly correlate with protected personal characteristics. \cite{mehrabiSurveyBiasFairness2021} identify several causes of unwanted bias including: \emph{i) historical bias} where the training data includes embedded historical social biases, \emph{ii) representation bias} where some groups with certain characteristics appear more frequently in the dataset, and \emph{iii) measurement bias} which arises from how certain features are measured or used.

Within the field of fair machine learning, several bias mitigation methods have been proposed, targeting the removal of unwanted bias from training data \citep[e.g.][]{zliobaiteHandlingConditionalDiscrimination2011,kamiranDataPreprocessingTechniques2012,%
feldmanCertifyingRemovingDisparate2015,iosifidisDealingBiasData2018,caldersBuildingClassifiersIndependency2009}, the training algorithm \citep[e.g.][]{huFairNNConjointLearning2020,iosifidisAdaFairCumulativeFairness2019,onetoTakingAdvantageMultitask2019,krasanakisAdaptiveSensitiveReweighting2018,%
kamiranDiscriminationAwareDecision2010,caldersThreeNaiveBayes2010,grariFairAdversarialGradient2019}, or the decisions made by a system \citep[e.g.][]{fishConfidenceBasedApproachBalancing2016,kamiranDecisionTheoryDiscriminationAware2012,kamiranDiscriminationAwareDecision2010,%
lohiaBiasMitigationPostprocessing2019}. These bias mitigation methods use different techniques and approaches and most rely on fairness metrics which are used to measure the amount of unwanted bias in a decision-making system. Additionally, the aim of bias mitigation methods is to optimise for these fairness metrics without decreasing the performance of a system \citep{fishConfidenceBasedApproachBalancing2016,zliobaiteRelationAccuracyFairness2015}.

In this paper, we
present a structured overview of bias mitigation methods for binary classification decision-making systems, analyse their benefits and limitations, summarise datasets as well as fairness and performance metrics used by these methods, and highlight directions for future work. Our contributions are as follows:

\begin{enumerate}
    \item We survey existing bias mitigation methods and report their limitations and benefits, datasets used, and the metrics used in evaluation.
    \item We review the limitations highlighted by existing methods and make recommendations for future bias mitigation methods for binary classification decision-making systems.
\end{enumerate}

The rest of the paper is organised as follows. In Section~\ref{background} we provide the necessary background, including types of bias mitigation methods, datasets, and metrics used, and  give an overview on existing, related surveys. In Section~\ref{methods}, we survey existing bias mitigation methods and compare their strengths and limitations. In Section~\ref{discussion} we provide recommendations for the development of future bias mitigation methods for binary classification decision-making systems based on the limitations identified, and conclude in Section~\ref{conclusion}.

\section{Background}
\label{background}

In this section, we present some key concepts, the different types of bias mitigation methods, as well as the main datasets, fairness and performance metrics used to evaluate the methods, and report on related work.

\subsection{Terminology}

Protected personal characteristics or simply \emph{protected attributes}, are features that define groups within populations which historically have different levels of privilege, but which should receive equal treatment \citep{bellamyAIFairness3602018,chakrabortyMakingFairML2020}.
These are often generalised as the features outlined in the US Equal Credit Opportunity Act
\footnote{\url{https://www.justice.gov/crt/equal-credit-opportunity-act-3}} and Fair Housing Act
\footnote{\url{https://www.justice.gov/crt/fair-housing-act-1}} such as race, sex, religion etc., but are specific to the context and application of the decision-making model being used.

\emph{Proxy attributes} are non-protected attributes that correspond to a protected attribute \citep{quySurveyDatasetsFairnessaware2021}. These attributes may correlate with a protected attribute and may result in reinforcing stereotypes. For example, a system that decides the cost of an individual's insurance should not make its decisions based on the protected attribute of ``race''; however it may make a decision partly based on the attributes ``address'' and ``smoker/non-smoker'' \citep{vannuenenTransparencyWhomAssessing2020} which may have strong correlations with the attribute ``race''.

Bias mitigation methods aim to ensure a system's decision does not discriminate based on the protected or proxy attributes. In the rest of the paper, we will refer to protected and proxy attributes as \emph{sensitive attributes}. An individual is said to be in the \emph{unprivileged group} if the value of their sensitive attribute defines them in the historically disadvantaged group.

\subsection{Types of Bias Mitigation Methods}

Bias mitigation methods are usually split into pre-processing, in-processing, and post-processing. In Section~\ref{methods}, we will present the approaches for the three types of bias mitigation methods

Pre-processing methods are used to mitigate bias in the training data, and are useful when we have access to the training data, but not necessarily to the machine learning model trained on the data \citep{bellamyAIFairness3602018}.

In-processing methods are used while training ML models, and are useful when we have access to the machine learning model to ensure that unwanted bias is not present in the output due to the training algorithm~\citep{zhangFAHTAdaptiveFairnessaware2019}. These methods can be model-agnostic or model-specific. Whilst model-agnostic methods may need to compromise on performance to be able to work for all models, model-specific methods are designed for a particular type of training algorithm, e.g. Naive Bayes \citep{caldersThreeNaiveBayes2010}, decision trees \citep{kamiranDiscriminationAwareDecision2010}, neural networks \citep{huFairNNConjointLearning2020}. These can be more effective as are designed specifically for the model chosen, but do not generalise.


Post-processing methods mitigate bias in the output of trained ML models. These methods focus on mitigating bias in the decision maker's output and are not concerned with the dataset or the training algorithm. Most are model-agnostic, however \cite{DBLP:conf/nips/SavaniWG20} recognises a subgroup named intra-processing methods which relies on some knowledge of the classifier, e.g. decision tree nodes \cite{kamiranDiscriminationAwareDecision2010}, posterior probabilities \cite{caldersThreeNaiveBayes2010}.

\subsection{Datasets} \label{datasets}

There are several datasets used to evaluate bias mitigation methods, which have been collected from real-world data. The datasets used in the methods collated are shown in Table \ref{tab:datasets}, ordered by the number of methods they are used for. We find that the most commonly used datasets in the literature are the \adult\ census 1994, \german\ credit, \compas\ and \dutch\ census datasets.

\newcommand{\source}[1]{\SetCell[c=4]{l} #1 & & &}
\definecolor{light-gray}{gray}{0.97}
\definecolor{medium-gray}{gray}{0.90}

\begin{table}[ht]
\centering
\small
\resizebox{\textwidth}{!}{
\begin{tabular}{l|l|r|r|r}

\textbf{Dataset} & \textbf{Abbreviation} & \textbf{\# instances} & \textbf{\#  attributes} & \textbf{\# methods} \\ \hline \hline
Adult Census 1994 \tablefootnote{\url{https://archive.ics.uci.edu/ml/datasets/Adult}} & \adult & 48842 & 15 & 32 \\ \hline
German Credit \tablefootnote{\url{https://archive.ics.uci.edu/ml/datasets/statlog+(german+credit+data)}} & \german & 1000 & 21 & 15 \\ \hline
COMPAS \tablefootnote{\url{https://www.kaggle.com/danofer/compass}} & \compas & 7214 & 51 & 9 \\ \hline
Dutch Census \tablefootnote{\url{https://microdata.worldbank.org/index.php/catalog/2101}} & \dutch & 189725 & 12 & 8 \\ \hline
Communities and Crime \tablefootnote{\url{https://archive.ics.uci.edu/ml/datasets/Communities+and+Crime}} & \communities & 1994 & 127 & 6 \\ \hline
Bank Marketing \tablefootnote{\url{https://archive.ics.uci.edu/ml/datasets/bank+marketing}} & \bank & 45211 & 17 & 6 \\ \hline
Heritage Health Prize \tablefootnote{\url{https://www.kaggle.com/c/hhp}} & \health & 147473 & 139 & 2 \\ \hline
Ricci vs DeStefano \tablefootnote{\url{https://www.openml.org/d/42665}} & \ricci & 118 & 6 & 2 \\ \hline
Credit Card Default \tablefootnote{\url{https://archive.ics.uci.edu/ml/datasets/default+of+credit+card+clients}} & \default & 30000 & 24 & 1 \\ \hline
KDD Census \tablefootnote{\url{https://archive.ics.uci.edu/ml/datasets/Census-Income+(KDD)}} & \kdd & 299285 & 41 & 1 \\ \\
\end{tabular}
}
\caption{Datasets used to evaluate bias mitigation methods and their number of instances and attributes, as well as the number of methods they are used in.}
\label{tab:datasets}
\end{table}

\subsection{Metrics} \label{metrics}
Bias mitigation methods are evaluated by calculating the amount of unwanted bias before and after the methods are applied; this can be achieved using fairness metrics. Fairness metrics can be split into two categories: \emph{group} and \emph{individual}. The aim of the methods is to decrease the level of unwanted bias according to a fairness metric without decreasing the performance of the model \citep{fishConfidenceBasedApproachBalancing2016,zliobaiteRelationAccuracyFairness2015}.

\paragraph{Group fairness metrics} measure unwanted bias across two groups defined by two values of a single personal characteristic (e.g. `male' and `female'). Table~\ref{tab:fairness-table} shows the group fairness metrics that are used in two or more of the collated methods including how they are calculated. These definitions are consistent with \cite{DBLP:journals/csur/PessachS23} and \cite{DBLP:conf/icse/VermaR18} which also provide further intuitive descriptions of the metrics. In the table, $A$ is an individual in the dataset, $\bar{y}$ is the classification for that individual, $y$ is the true label of the individual in the training data, and $s$ is the unprivileged group. Hence, $A \in{s}$ means that the individual $A$ is in the unprivileged group; $\bar{y}=1$ ($=0$) means that the classifier predicts the outcome $1$ ($0$), respectively; and $y=1$ ($=0$) means that the true label of the individual is $1$ ($0$), respectively.
Other metrics used in the bias mitigation methods collated that are not included in Table~\ref{tab:fairness-table} are discussed in Section~\ref{methods}.

\begin{table}[ht]
\centering
\small
\begin{tabular}{l|l}

\textbf{Metric}             & \textbf{Calculation}\\ \hline \hline
Equal opportunity &  $|P(\bar{y}=1|A \notin{s},y=1) - P(\bar{y}=1|A \in{s},y=1)|$ \\ \hline

Equalised odds &  $|P(\bar{y}=1|A \notin{s},y) - P(\bar{y}=1|A \in{s},y)|, y\in \lbrace 0,1 \rbrace$ \\ \hline

Demographic parity & $|P(\bar{y}=1|A \notin{s}) - P(\bar{y}=1|A \in{s})|$ \\ \hline

Disparate impact & $|P(\bar{y}=1|A \notin{s}) / P(\bar{y}=1|A \in{s})|$ \\ \hline

Disparate mistreatment & $|P(\bar{y}=0|A \notin{s},y=1) - P(\bar{y}=0|A \in{s},y=1)| + $ \\ & $|P(\bar{y}=1|A \notin{s},y=0) - P(\bar{y}=1|A \in{s},y=0)|$ \\

\end{tabular}
\caption{Commonly used group fairness metrics.}
\label{tab:fairness-table}
\end{table}

The metrics are calculated using the values from the confusion matrix (true positives (TP), false positives (FP), true negatives (TN), and false negatives (FN)) across two values of the sensitive attribute. For a binary decision-making system, TP are the correct positive classifications ($\bar{y} = 1, y = 1$), FP are the incorrect positive classifications ($\bar{y} = 1, y=0$), TN are the correct negative classifications ($\bar{y}=0,y=0$) and FN are the incorrect negative classifications ($\bar{y}=0,y=1$). The rates of these measures are defined below. \\

\noindent \emph{True positive rate (TPR)} $\displaystyle = \frac{TP}{TP+FN}$, TP out of total actual positive classifications\\ \\
\emph{False positive rate (FPR)} $\displaystyle = \frac{FP}{TN+FP}$, FP out of total actual negative classifications.\\ \\
\emph{True negative rate (TNR)} $\displaystyle = \frac{TN}{TN+FP}$, TN out of total actual negative classifications.\\ \\
\emph{False negative rate (FNR)} $\displaystyle = \frac{FN}{TP+FN}$, FN out of total actual positive classifications.\\ \\

From Table~\ref{tab:fairness-table}, $P(\bar{y}=1|A \in{s},y=1)$ represents the TPR for the unprivileged group and $P(\bar{y}=1|A \notin{s},y=1)$ represents the TPR for the privileged group.

\emph{Equal opportunity} \citep{hardtEqualityOpportunitySupervised2016} therefore aims to have equal TPRs for the privileged and unprivileged group. For example, when predicting fraudulent activity on a bank account, the primary aim is to correctly identify fraudulent activity (TP). It is important that the unprivileged group does not receive fewer TP as this would put them at a disadvantage.

\emph{Equalised odds} \citep{hardtEqualityOpportunitySupervised2016} places an emphasis on minimising the difference in TPRs and FPRs across the privileged and unprivileged groups as represented by $P(\bar{y}=1|A \notin{s},y) - P(\bar{y}=1|A \in{s},y)$, for both $y=1$ and $y=0$. The TPR is represented when $y=1$ and the FPR is represented when $y=0$. Bias mitigation methods that use this metric aim to minimise these separately and then sum or average the differences for a single metric value. An example of where this metric may be suitable is when predicting who will be able to pay back a loan --- it is important to ensure fairness for individuals who are correctly predicted to pay back the loan as well as incorrectly, as this could lead to negative consequences for individuals in the future such as damage to a credit score and difficulty getting another loan in the future.

\emph{Demographic parity} and \emph{disparate impact} are concerned with achieving the same rates of positive classifications across the privileged and unprivileged groups. This can be used to improve diversity with respect to an identified group, for example improving demographic parity in a hiring system to increase the number of women selected \citep{realLifeFairnessNotions}.

\emph{Disparate mistreatment} \citep{DBLP:conf/www/ZafarVGG17} combines TPR, FPR, TNR, and FNR to give an overall view of the system;  this is useful when concerned with FNRs, for example in healthcare where a false negative could be disastrous to an individual if a condition is not predicted early.

\paragraph{Individual fairness metrics} have been created to evaluate the unwanted bias of an individual decision. \cite{dworkFairnessAwareness2012} created an individual fairness metric, Dwork IF, which measures to what extent similar individuals receive the same classification, where the definition of a similar individual is required as input and depends on the context. The metric quantifies the number of pairs of similar individuals receiving the same classification.

\cite{zemelLearningFairRepresentations2013} introduced consistency, an individual fairness metric which defines the similarity of individuals by considering its nearest neighbours according to Manhattan distance and providing a score based on the number of neighbours with different classifications. Consistency is defined below, where $n$ is the number of individuals in the dataset, $k$ is the number of neighbours chosen, $kNN(x_i)$ represents the k-nearest neighbours and $\hat{y}$ is an individual's label. $$\displaystyle 1 - \frac{1}{n} \sum_{i=1}^{n}|\hat{y}-\frac{1}{k}\sum_{j \in kNN(x_i)}\hat{y}|$$

\paragraph{Performance metrics} are used to evaluate the performance of decision-making systems. The metric values often decrease after the application of a bias mitigation method. The metrics used for the evaluation of the methods collated in this survey include {\em Accuracy}, {\em $F_1$}, and the {\em Balanced Error Rate} ($BER$), defined as follows.

\begin{eqnarray*}
Accuracy & = &\frac{TP +TN}{TP+TN+FP+FN}\\[1.5ex]
F_{1} & = &\frac{TP}{TP+ \frac{1}{2}(FP + FN)}\\[1.5ex]
BER & = & \frac{1}{2}\left(\frac{FN}{FN + TP}+\frac{FP}{FP + TN}\right)
\end{eqnarray*}

\subsection{Related Work}
\label{related}

In recent years, there has been a growing interest in the development of bias mitigation methods, and as a result, there have been several surveys conducted. In this section, we will review them and their contributions.

\cite{DBLP:journals/csur/PessachS23} is the most similar to our survey and provides a detailed background to bias mitigation, including metrics and datasets used throughout the literature. They also split the methods into pre-, in- and post-processing and include tables of methods collated, including a brief description of each method. Our survey differs by providing a novel contribution in finding the limitations of each surveyed method, and using these limitations to provide evidence for recommendations for future directions of research. We hope this will be able to guide researchers into a new direction for creating bias mitigation methods.

\cite{Dunkelau2020FairnessAwareML} gives an extensive overview of fairness in machine learning, overlapping with our survey and that of \cite{DBLP:journals/csur/PessachS23} in their descriptions of datasets and metrics. They describe the methods, but do not provide an analysis or critique of them as our survey does.
\cite{hortBiasMitigationMachine2022} present a comprehensive survey of bias mitigation methods for machine learning which categorises methods from 234 papers into pre-, in-, and post-processing, and  further by the approach they take, e.g. sampling, adversarial, adding constraints. Furthermore, they provide statistics about the methods in terms of datasets and metrics used. Some of the methods they survey however are small extensions of other methods
or not from peer-reviewed work. Our survey differs from these works by providing a discussion of bias mitigation methods, specifically, commenting on their limitations which should be considered for the development of new methods.

Other related work focuses on the development of frameworks to be able to more easily evaluate bias mitigation methods on different models and datasets, and using different fairness metrics. These are outlined in \cite{Dunkelau2020FairnessAwareML}. Along with frameworks, quantitative analyses of several methods are usually provided, but the usability or limitations of the methods are not commented on. For example, \cite{friedlerComparativeStudyFairnessenhancing2019} developed a framework that allows the comparison of different methods on the same dataset and model, for different fairness metrics. In addition, \cite{schelterFairPrepPromotingData2019} designed FairPrep which allows users to tune hyper-parameters of  classifiers to see the impact on fairness and accuracy metrics across several datasets (i.e. Adult, German Credit and Ricci vs DeStefano).

\section{Current State of Research}
\label{methods}

In this section, we survey existing bias mitigation methods for binary classification and compare their strengths and limitations. We split them into pre-, in-, and post-processing and describe the method, datasets and metrics used to evaluate them.

Other methods not designed for binary classification, which are not included in Sections \ref{subsec:preprocess}--\ref{subsec:postprocess}, include ones designed to improve fairness in regression \citep{caldersControllingAttributeEffect2013,berkConvexFrameworkFair2017,chzhenFairRegressionPlugIn2020}, multi-class classification \citep{putzelBlackboxPostProcessingMulticlass2022,alghamdiAdultCOMPASFairness2022}, clustering \citep{chierichettiFairClusteringFairlets2017,zikoVariationalFairClustering2021,abbasiFairClusteringEquitable2021,backursScalableFairClustering2019,rosnerPrivacyPreservingClustering2018,beraFairAlgorithmsClustering2019}, outlier detection \citep{pFairOutlierDetection2020}, clustering from demonstrations \citep{galhotraLearningGenerateFair2021}, in online data streams \citep{iosifidisFABBOOOnlineFairnessaware,zhangFAHTAdaptiveFairnessaware2019} and where there is little data available \cite{slackFairnessWarningsFairMAML2019}. These methods are evaluated on the datasets identified in the fairness domain but are not comparable to the other methods surveyed in Tables~\ref{tab:pre-table}--\ref{tab:post-table}.

The methods collated in the Appendix of \cite{quySurveyDatasetsFairnessaware2021} are the basis for conducting our survey of bias mitigation methods. These methods were obtained by a systematic search process, which involved the use of Google Scholar to find all papers that used fairness datasets for evaluation, in conjunction with the search terms ``bias'', ``discrimination'' and ``fairness''. The resulting papers were then filtered based on the number of citations as well as the publication venue (i.e. conferences and journals).
The filtering process ensured that only papers of high quality and relevance were included in the survey, providing a reliable representation of the state-of-the-art in the field. Additionally, we filter the papers by those that propose bias mitigation methods for binary classifiers and find 25 prominent papers from which we extract 33 bias mitigation methods to analyse.

A limitation of the bias mitigation methods collated is that they cannot be directly compared due to being evaluated using different fairness metrics and different models trained on datasets of different sizes and distributions.

\subsection{Pre-processing Methods}
\label{subsec:preprocess}

Pre-processing methods focus on removing unwanted bias from the training data. By definition, this means that all these methods are model-agnostic as they are not concerned with the machine learning model used. A limitation of pre-processing methods is that they are considered intrusive because they change the dataset \citep{kamiranDataPreprocessingTechniques2012}. Next, we review the pre-processing methods in the literature and summarise the datasets and metrics used by these methods in Table~\ref{tab:pre-table}.

\begin{table}[ht]
\begin{adjustbox}{width=1\textwidth}
\footnotesize
    \begin{tabular}{lllll}
        \textbf{Method} & \textbf{Source}  & \textbf{Datasets} & \textbf{Fairness Metrics} & \textbf{Performance} \\
         & & & & \textbf{Metrics} \\
        \hline
        \hline

        Suppression & \cite{kamiranDataPreprocessingTechniques2012} & \adult, \german, \dutch, & Equal opportunity & Accuracy \\
        & & \communities & & \\
        \hline
        Massaging & \cite{kamiranDataPreprocessingTechniques2012} & \adult, \german, \dutch &  Equal opportunity &  Accuracy \\
        \hline
        Reweighing & \cite{kamiranDataPreprocessingTechniques2012} & \adult, \german, \dutch &  Equal opportunity &  Accuracy \\
        \hline
        Uniform Sampling & \cite{kamiranClassificationNoDiscrimination2010,kamiranDataPreprocessingTechniques2012} & \adult, \german, \dutch & Equal opportunity &  Accuracy \\
        \hline
        Preferential Sampling & \cite{kamiranClassificationNoDiscrimination2010,kamiranDataPreprocessingTechniques2012} & \adult, \german, \dutch & Equal opportunity &  Accuracy \\
        \hline
        Local Massaging & \cite{zliobaiteHandlingConditionalDiscrimination2011} & \adult, \dutch, & Illegal  & Accuracy \\
        & \cite{kamiranQuantifyingExplainableDiscrimination2013} & \communities &  discrimination & \\
        \hline
        Local Preferential & \cite{zliobaiteHandlingConditionalDiscrimination2011} &  \adult, \dutch, & Illegal &  Accuracy \\
        Sampling & \cite{kamiranQuantifyingExplainableDiscrimination2013} & \communities & discrimination & \\
        \hline
        Oversampling & \cite{iosifidisDealingBiasData2018} &  \adult, \german & Equalised odds & $F_{1}$ \\
        \hline
        SMOTE-approach & \cite{iosifidisDealingBiasData2018} & \adult, \german & Equalised odds & $F_{1}$\\
        \hline
        Geometric & \cite{feldmanCertifyingRemovingDisparate2015} &  \adult, \german, &  Disparate impact, & Accuracy, \\
        Repair & \cite{feldmanComputationalFairnessPreventing2015} & \ricci & Demographic parity & BER\\
        \hline
        Combinatorial & \cite{feldmanCertifyingRemovingDisparate2015} &  \adult, \german, &  Disparate impact, & Accuracy, \\
        Repair & \cite{feldmanComputationalFairnessPreventing2015} & \ricci & Demographic parity & BER \\
        \hline
        Learning Fair & \cite{zemelLearningFairRepresentations2013} &  \adult, \german, \health & Demographic parity, & Accuracy \\
        Representations & & & Consistency & \\
        \hline
        LCIFR & \cite{ruossLearningCertifiedIndividually2020} &  \adult, \german, &  Certified individual & Accuracy \\
        & & \compas, \health, & fairness & \\
        & & \communities & & \\ \\

    \end{tabular}
    \end{adjustbox}
    \caption{Table of collated pre-processing bias mitigation methods.}

    \label{tab:pre-table}
\end{table}

\paragraph{\cite{kamiranClassificationNoDiscrimination2010,kamiranDataPreprocessingTechniques2012}} proposed five pre-processing methods, namely \textit{Suppression}, \textit{Massaging}, \textit{Reweighing}, \textit{Uniform Sampling},  and \textit{Preferential Sampling}.
\textit{Suppression} simply removes the protected attribute and is often used as a baseline. However, this approach is problematic as there may still be proxy attributes which may lead to a biased model.
\textit{Massaging} changes the labels of some individuals in the training set chosen based on a learned ranker that selects individuals close to the decision boundary to be `promoted' (i.e. from $-$ to $+$) and `demoted' (i.e. from $+$ to $-$). The number of individuals whose labels are changed are selected based on the calculated level of unwanted bias according to the equal opportunity fairness metric.

\textit{Reweighing} assigns weights to individuals based on the associated classification and whether they are in the unprivileged group. In particular, four weights are calculated for the following combinations: positive/negative classification (not) in the unprivileged group.
For example, all individuals with a positive classification in the unprivileged group will be assigned the same weight which will be greater than the weight assigned to the individuals with a negative classification in the unprivileged group.
Reweighing can be applied on its own when the training algorithm uses weights (e.g. linear-based classifiers), otherwise we need to \emph{sample} individuals from the dataset according to these weights.
\textit{Uniform Sampling} constructs a dataset by choosing how many individuals to duplicate or remove according to their weight in each of the four groups mentioned above, and selects which individuals within the group to choose with uniform probability.
\textit{Preferential Sampling} similarly constructs a dataset by choosing how many individuals in each of the four combinations to duplicate and remove by their weight. However, it uses a ranker to decide which individuals are close to the decision boundary between a positive and negative classification and prioritises these individuals to duplicate or remove.


\noindent \textbf{Limitation:} The methods described above only allow for a single binary sensitive attribute in the dataset. The definitions of existing group fairness metrics (found in Table~\ref{tab:fairness-table}) only account for a single binary sensitive attribute meaning methods which use these metrics only allow the removal of unwanted bias with respect to that one attribute. Group fairness metrics would need to be adapted to account for multiple binary sensitive attributes and sensitive attributes with multiple values.\footnote{Unless otherwise stated, all methods discussed in this survey that use these group fairness metrics only consider a single binary sensitive attribute.}

\noindent \textbf{Datasets:} All methods are evaluated on \adult, \communities, and \german\ datasets.

\noindent \textbf{Fairness metrics:} Equal opportunity.

\paragraph{\cite{kamiranQuantifyingExplainableDiscrimination2013,zliobaiteHandlingConditionalDiscrimination2011}} extended the work by \cite{kamiranDataPreprocessingTechniques2012} and introduced two bias mitigation methods, \textit{Local Massaging} and \textit{Local Preferential Sampling}. They propose a new notion for measuring the fairness of a dataset called \textit{illegal discrimination}, where some bias can be explained and is therefore acceptable to be included in the model, while other types of bias are unwanted (i.e. `illegal').

Illegal discrimination makes use of explanatory attributes, which are attributes that correlate to a sensitive attribute but should still be used in making the decision. Determining explanatory attributes is domain dependent. For example, assuming that ``gender'' is the sensitive attribute in a decision-making system for monthly salaries, the correlated attribute of ``relationship'' (e.g. husband or wife) would not be an explanatory attribute as there should not be decisions based on this attribute. However, the attribute ``working hours'', which may also be correlated to gender, could be an explanatory attribute as it may be informative in determining monthly salaries.

Illegal discrimination is measured as the difference between the level of unwanted bias according to an existing group fairness metric and the explained bias as shown below, where $k$ is the number of explanatory attributes, $e_{i}$ is a value of the explanatory attribute values, and $A$ and $B$ are the values of the sensitive attribute (e.g. male, female):


\[
 D_{expl} = \sum_{i=1}^{k}(P(e_{i}|A) - P(e_{i}|B)) P^{*}(+|e_{i}) \qquad P^{*}(+|e_{i}) = \frac{P(+|e_{i}, A) + P(+|e_{i},B)}{2}
\]

\noindent $P(e_{i}|A)$ ($P(e_{i}|B)$) is the probability that an individual has the explanatory attribute value $e_{i}$ given they are in the unprivileged group (in the privileged group), respectively; and $P^{*}(+|e_{i})$ represents the probability of an individual having a positive classification given it has the explanatory attribute value $e_{i}$.

\medskip\noindent \textbf{Limitations:}
\begin{enumerate}
    \item

    The choice of datasets for evaluating the two bias mitigation methods impacts the results, making it difficult to compare and assess the usefulness of the proposed methods.
    The methods are shown to be more effective when the explanatory attribute is highly correlated with the sensitive attribute. In particular, the explanatory attribute has a higher correlation with the sensitive attribute in the \adult\ dataset than in the \communities\ and \dutch\ datasets, resulting in a lower level of illegal discrimination on the \adult\ dataset than on the other datasets.

    \item An issue with using the illegal discrimination metric is that the explanatory attributes have to be decided by a domain expert. This raises questions about whom the domain expert should be and whether there is potential for human bias or error in choosing these attributes.
\end{enumerate}

\noindent \textbf{Datasets:} The methods are evaluated on \adult, \communities, and \dutch\ datasets.

\noindent \textbf{Fairness metrics:} Illegal discrimination.

\paragraph{\cite{iosifidisDealingBiasData2018}} introduced two pre-processing methods. \textit{Oversampling} is a naive method which selects random individuals in the unprivileged group and duplicates them to ensure a balanced training dataset. They also propose using \textit{SMOTE} \citep{DBLP:journals/jair/ChawlaBHK02} to create new individuals in the training data for each individual in the unprivileged group. The method finds the k-nearest neighbours for each individual that are also in the unprivileged group. It chooses randomly amongst these neighbours and generates new individuals in the unprivileged group that lie on the line between the individual and the chosen neighbour.


\noindent \textbf{Limitation:} There is a trade-off between fairness and performance as the performance was shown to decrease after applying each bias mitigation method. There is no discussion as to whether this performance decrease is acceptable.

\noindent \textbf{Datasets:} The methods are evaluated on the \adult\ and \german\ datasets.

\noindent \textbf{Fairness metrics:} Equalised odds.

\paragraph{\cite{feldmanCertifyingRemovingDisparate2015,feldmanComputationalFairnessPreventing2015}} proposed two methods, \textit{Geometric Repair} and \textit{Combinatorial Repair},  which ensure that non-sensitive attributes are not used to predict stratifying columns, where stratifying columns are sensitive attributes or combinations of sensitive attributes (e.g. white-male). This is achieved by changing the values of non-stratifying columns to ensure the distributions over the different values of the stratifying columns are the same (e.g. white-males and white-females have the same median values for every other attribute). Geometric Repair and Combinatorial Repair include a parameter which allows the specification of partial repair, corresponding to the desired trade-off of performance and fairness. The methods are evaluated using disparate impact and demographic parity which are consistent with the notion of fairness used throughout law and regulation \citep{feldmanComputationalFairnessPreventing2015}. One basis for assessing whether a decision-maker is legally fair is the 80\% rule which is a minimum level of disparate impact that should be accepted \citep{agarwalReductionsApproachFair2018}. In other terms, the system is unfair if the disparate impact metric calculated is less than 80\% as below:


\[
  \frac{\textrm{Rate of individuals in the unprivileged group receiving a positive outcome}}{\textrm{Rate of individuals in the privileged group receiving a negative outcome}} < 0.8
\]

\noindent \textbf{Benefits:}
\begin{enumerate}
    \item The methods can be applied to datasets with multiple sensitive attributes as the authors propose stratifying attributes which combine all values of the sensitive attributes, e.g. white-male, black-female.
    \item Different levels of disparate impact can be removed using both Geometric Repair and Combinatorial Repair. The partial repair of the dataset is achieved by including a parameter $\lambda$ in the methods, which is a value between 0 and 1, that specifies to what extent the data is `repaired'.
\end{enumerate}

\noindent \textbf{Limitation:} Enforcing the median values of attributes to be the same across stratifying attributes requires the attributes to be orderable. Attributes can be categorical only if categories have a natural order.

\noindent \textbf{Datasets:} The methods are evaluated on \adult, \german, and \ricci\ datasets.

\noindent \textbf{Fairness metrics:} Disparate impact.

\paragraph{\cite{zemelLearningFairRepresentations2013}} introduced \textit{Learning Fair Representations (LFR)} which aims to solve an optimisation problem to optimise for group fairness, individual fairness and performance, by encoding accuracy, demographic parity and  their created individual fairness metric of \emph{consistency} as objective functions. Consistency compares an individual to its k-nearest neighbours and counts how many of the neighbours' labels differ.

\noindent \textbf{Limitation:} The Learning Fair Representations method optimises for demographic parity and consistency and there are no parameters to allow for the partial optimisation of either metric. This may not always be desired as it could cause positive discrimination or lead to a significant decrease in performance.

\noindent \textbf{Datasets:} The method is evaluated on the \adult, \german, and \health\ datasets.

\noindent \textbf{Fairness metrics:} Demographic parity, Consistency.

\paragraph{\cite{ruossLearningCertifiedIndividually2020}} proposed \textit{Learning Certified Individually Fair Representations} \textit{(LCIFR)} which modifies the dataset using a neural network with a loss function based on a user-specified individual similarity definition. The authors propose \emph{certified individual fairness}, a metric which calculates the proportion of individuals with the same label given all similar individuals.

\noindent \textbf{Limitation:} The method does not ensure that similar individuals are not treated negatively as it only targets improving individual fairness. For example, individuals with the attribute value female may be considered similar and receive the same classification, therefore satisfying individual fairness. However, if these classifications are all negative, then group fairness with respect to gender would not be satisfied.

\noindent \textbf{Datasets:} The method is evaluated on the
\adult, \compas, \communities, \german, and \health\ datasets.

\noindent \textbf{Fairness metrics:} Certified individual fairness.

\subsection{In-processing Methods} \label{subsec:inprocess}

In-processing methods can be model-agnostic or model-specific and usually involve adjusting the training algorithm, training adversarial learners or optimising weights for both performance and fairness. As with pre-processing methods, in-processing methods are evaluated/optimised for a range of different fairness and performance metrics, as well as being tested on a range of different datasets. The methods collated are shown in Table \ref{tab:in-table}, along with the classifier, datasets, and metrics used.

\begin{table}[ht]
\begin{adjustbox}{width=1\textwidth}
    \begin{tabular}{llllll}

        \textbf{Method} & \textbf{Source} & \textbf{Classifier} & \textbf{Datasets} & \textbf{Fairness Metrics} & \textbf{Performance} \\
        & & & & & \textbf{Metrics} \\
        \hline
        \hline
        Fair Learner for  & \cite{choiLearningFairNaive2020} & Naive Bayes & \adult, & Own discrimination & Accuracy \\
        Naive Bayes & & & \german, & score (Equation \ref{eq:discscore}) & \\
        & & & \compas & & \\
        \hline
        Two Naive Bayes Model & \cite{caldersThreeNaiveBayes2010} & Naive Bayes & \adult & Disparate impact & Accuracy \\
        \hline
        Latent Variable Model & \cite{caldersThreeNaiveBayes2010} & Naive Bayes & \adult & Disparate impact & Accuracy \\
        \hline
        Bayesian Network  & \cite{mancuhanCombatingDiscriminationUsing2014} & Bayesian  & \adult, & Count of discriminated & Accuracy \\
        Learning & & Network & \german & instances & \\
        \hline
        Fair NN & \cite{huFairNNConjointLearning2020} & Neural  & \adult,  & Equalised odds & Accuracy, \\
        & & Network & \bank & & BER \\
        \hline
        Decision Tree Learning & \cite{kamiranDiscriminationAwareDecision2010} & Decision Tree & \adult,  & Disparate impact & Accuracy \\
        & & & \dutch & & \\
        \hline
        Regularization-Inspired & \cite{bechavodLearningFairClassifiers2017} & Logistic & \compas, & FPR, FNR & Accuracy \\
        Approach & & Regression & \default & & \\
        \hline
        Adversarial Learning & \cite{zhangMitigatingUnwantedBiases2018} & Logistic & \adult & Equalised odds, & FPR, FNR \\
        & & Regression & & Equal opportunity, &  \\
        & & & & Demographic parity & \\
        \hline
        Fair Adversarial   & \cite{grariFairAdversarialGradient2019} & Gradient Tree  & \adult,  & Equalised odds, & Accuracy \\
         Gradient Tree Boosting & & Boosting & \bank, & Demographic parity & \\
        & & & \compas & & \\
        \hline
        Mechanisms for Fair & \cite{zafarFairnessConstraintsMechanisms2017} & Convex  & \adult,  & Disparate impact, & Accuracy \\
        Classification & & margin-based & \bank &  Disparate mistreatment & \\
        \hline
        AdaFair & \cite{iosifidisAdaFairCumulativeFairness2019} & Model-agnostic & \adult, & Equalised odds & BER \\
        & & & \bank,  & & \\
        & & & \compas & & \\
        & & & \kdd & & \\
        \hline
        Adaptive Sensitive & \cite{krasanakisAdaptiveSensitiveReweighting2018} & Model-agnostic & \adult, & Disparate impact,
        & Accuracy \\
        Reweighting & & & \bank, & & \\
        & & & \compas & & \\
        \hline
        Privileged Learning \& & \cite{quadriantoRecyclingPrivilegedLearning2017} & Model-agnostic & \adult, & Equalised odds, & Accuracy \\
         Distribution Matching & & & \compas & Equal opportunity, & \\
        & & & & Disparate mistreatment, & \\
        & & & & Demographic parity & \\
        \hline
        Fair Multitask Learning & \cite{onetoTakingAdvantageMultitask2019} & Model-agnostic & \adult, & Equalised odds, & Accuracy \\
        & & & \compas & Equal opportunity & \\ \\

    \end{tabular}
    \end{adjustbox}
    \caption{Table of collated in-processing bias mitigation methods.}

    \label{tab:in-table}
\end{table}

\paragraph{\cite{choiLearningFairNaive2020}} introduced \textit{Fair Learner for Naive Bayes} which proposes discrimination patterns that are defined as individuals that have different classifications depending on whether the sensitive attribute has an impact on the decision (i.e. if the sensitive attribute is removed and the classification changes, there is a discrimination pattern). The aim of the method is to discover these discrimination patterns using a branch-and-bound method \citep{morrisonBranchandboundAlgorithmsSurvey2016} and iteratively learn a Naive Bayes classifier that minimises a discrimination score based on the probability of sensitive attributes affecting a positive classification. The discrimination score is defined in Equation \ref{eq:discscore}, where $P(d|xy)$ is the probability of a positive classification given all the attributes, calculated as the proportion of positive classifications when the classifier is trained with all attributes. $P(d|y)$ is the probability of a positive classification given all attributes except the sensitive attribute, calculated as the proportion of positive classifications when the classifier is trained with all attributes except the sensitive attribute.
\begin{eq}
$$\Delta = P(d|xy) - P(d|y)$$
\label{eq:discscore}
\end{eq}

\noindent \textbf{Benefit:} This method can be applied to datasets with multiple sensitive attributes.

\noindent \textbf{Limitations:}
\begin{enumerate}
\item A new fairness metric is created, but a comparison with existing metrics is not provided. However, the authors justify the use of the new metric by observing that current common fairness metrics consider only one sensitive attribute.
\item The method can only be applied to Naive Bayes models.
\end{enumerate}

\noindent \textbf{Datasets:} The method is applied on \adult, \german, and \compas\ datasets.

\noindent \textbf{Fairness metrics:} Own discrimination score (Equation \ref{eq:discscore}).

\paragraph{\cite{caldersThreeNaiveBayes2010}} introduced two in-processing methods for Naive Bayes classifiers: \textit{Latent Variable Model} and \textit{Two Naive Bayes Model}.
\textit{Latent Variable Model} aims to classify individuals given a discrimination-free model. This involves modelling classifications as hidden variables and using an expectation maximisation approach to find the individuals in the negative class who should be in the positive class and vice-versa, and swapping their classifications.
The \textit{Two Naive Bayes} method learns two different models for the data, split by the values of the identified sensitive attribute, and uses these to classify individuals, using the model corresponding to their value of the sensitive attribute. This method was shown to yield the best results.

\noindent \textbf{Limitations:}
\begin{enumerate}
    \item The authors compare their Two Naive Bayes method to Fair Learner for Naive Bayes \citep{choiLearningFairNaive2020} on the \adult\ dataset and find that it yields worse accuracy. However, it is difficult to directly compare the impact on fairness due to the different metrics used (i.e. own discrimination score (Equation \ref{eq:discscore}) versus disparate impact).
    \item The methods can only be applied to Naive Bayes models.
\end{enumerate}

\noindent \textbf{Datasets:} The methods are evaluated on the \adult\ dataset.

\noindent \textbf{Fairness metrics:} Disparate impact.

\paragraph{\cite{mancuhanCombatingDiscriminationUsing2014}} proposed \textit{Bayesian Network Learning} which learns a Bayesian network and then
creates a new network by removing the sensitive attributes and the parents and children of the sensitive attributes. A discriminated instance is an individual whose classification in the new network differs from its classification in the original network. The labels of these discriminated instances are then swapped and the aim of the bias mitigation method is to learn a new Bayesian network based on the updated dataset. The number of discriminated instances is used as the metric to measure the level of unwanted bias.

\noindent \textbf{Benefit:} The method does not use the sensitive attribute at test time. Whilst it uses the sensitive attribute to identify bias and then mitigate it, the method does not use sensitive attributes to make predictions.

\noindent \textbf{Limitations:}
\begin{enumerate}
    \item A new metric (i.e. number of discriminated instances) is created for evaluation, which does not encapsulate the complexity of measuring fairness.
    \item This method can only be applied to Bayesian networks.
\end{enumerate}

\noindent \textbf{Datasets:} The method is tested on the \adult\ and \german\ datasets.

\noindent \textbf{Fairness metrics:} Number of discriminated instances.

\paragraph{\cite{huFairNNConjointLearning2020}} proposed \textit{Fair NN} which improves fairness in neural networks by altering the network's autoencoder with the aim to suppress sensitive attributes and constrain the loss function based on the fairness metric of equalised odds.

\noindent \textbf{Limitations:}
\begin{enumerate}
    \item As with most of the methods surveyed, this method only allows for a single binary sensitive attribute. \cite{huFairNNConjointLearning2020} proposed a mechanism to account for a multi-value sensitive attribute by pre-processing the dataset to only have binary values. However, this would result in multiple sensitive attributes, which cannot be addressed with the current method.
    \item The method can only be applied to Neural Networks.
\end{enumerate}

\noindent \textbf{Datasets:} The method is evaluated on \adult\ and \bank\ datasets.

\noindent \textbf{Fairness metrics:} Equalised odds.

\paragraph{\cite{krasanakisAdaptiveSensitiveReweighting2018}} introduced \textit{Adaptive Sensitive Reweighting} which iteratively trains a classifier, each time adapting the training data to converge towards the best performance-fairness trade-off according to accuracy and disparate impact. At each iteration, weights for individuals are calculated based on the classification error and whether they are in the unprivileged group. For example, a misclassified individual in the unprivileged group will be given a different weight compared to a misclassified individual in the privileged group.



\noindent \textbf{Limitation:} Adaptive Sensitive Reweighting can greatly increase computational cost as it involves training the classifier until convergence, thus highlighting scalability issues.

\noindent \textbf{Datasets:} The method is evaluated on \adult, \bank, and \compas\  datasets.

\noindent \textbf{Fairness metrics:} Disparate impact, Disparate mistreatment.

\paragraph{\cite{kamiranDiscriminationAwareDecision2010}} proposed two methods \textit{Discrimination-Aware Tree Construction} and \textit{Relabeling Decision Trees}. Discrimination-Aware Tree Construction changes the splitting criteria of branches by including the notion of fairness. Specifically, instead of splitting based on the information gain with respect to the class, the split is based on the information gain with respect to the class and the sensitive attribute.

\noindent \textbf{Benefit:} There is no detriment to the performance of the system by using this method. The results show a noticeable reduction in the disparate impact score without a loss in accuracy on the datasets evaluated.

\noindent \textbf{Limitations:}
\begin{enumerate}
    \item The initial level of unwanted bias in the training and test datasets can greatly impact the fairness metric values calculated and therefore impact the evaluation of new methods. The accuracy of the mitigated classifier increases over a standard decision tree learning classifier when there is unwanted bias in the training data but not in the test data.
    \item This method can only be applied to Decision Tree models.
\end{enumerate}

\noindent \textbf{Datasets:} The method is evaluated on \adult\ and \dutch\ datasets.

\noindent \textbf{Fairness metrics:} Disparate impact.


\paragraph{\cite{bechavodLearningFairClassifiers2017}} proposed a \textit{Regularization-Inspired Approach} that includes a fairness penalty into the loss function of a logistic regression model which penalises for different FPRs and FNRs across different values of the sensitive attribute.

\noindent \textbf{Benefit:} The method can be applied to datasets with one or two sensitive attributes. The method can almost entirely remove unwanted bias on the \compas\ dataset, given the difference of FPRs and FNRs with respect to two sensitive attributes, with results showing only a small reduction in accuracy.

\noindent \textbf{Limitations:} \begin{enumerate}
    \item The method does not provide theoretical guarantees that balancing the FPRs and FNRs across protected groups improves fairness in all cases.
    \item The method can be applied only to logistic regression models.
\end{enumerate}

\noindent \textbf{Datasets:} The method is evaluated on \compas\ and \default\ datasets.

\noindent \textbf{Fairness metrics:} False positive rates, False negative rates.

\paragraph{\cite{zhangMitigatingUnwantedBiases2018}} introduced an \textit{Adversarial Learning} method which aims to maximise the prediction accuracy of a classifier while minimising an adversarial classifier's ability to predict the sensitive attribute. Logistic regression is used for both classifiers, with the input to the adversarial model being the output from the classifier.


\noindent \textbf{Benefit:} The method allows the user to choose whether to optimise for equalised odds, equal opportunity, or demographic parity. This means that the method can be applied in different settings by specifying the desired notion of fairness, depending on context.

\noindent \textbf{Limitations:}
\begin{enumerate}
    \item The method relies on hyper-parameters that impact how well the method reduces the level of unwanted bias. In addition, more complex prediction problems (e.g. multi-class classifiers) may require more complex adversaries to be adapted.
    \item The method can only be applied to logistic regression models.
\end{enumerate}

\noindent \textbf{Datasets:} The method is evaluated on the \adult\ dataset.

\noindent \textbf{Fairness metrics:} Equalised odds, Equal opportunity, Demographic parity.

\paragraph{\cite{zafarFairnessConstraintsMechanisms2017}} proposed \textit{Mechanisms For Fair Classification} which defines a ``novel intuitive measure of decision boundary (un)fairness'' that measures the variance of the decision boundary and uses it to formulate an approach that either aims to maximise fairness with accuracy constraints or aims to maximise accuracy with fairness constraints, depending on the preference of the user of the system towards prioritising fairness or performance.

\noindent \textbf{Benefit:} The method can be applied to datasets with multiple sensitive attributes and multi-value sensitive attributes.

\noindent \textbf{Limitation:} The authors do not provide theoretical guarantees that the method decreases the level of unwanted bias in any system, for any dataset. Providing guarantees that the constraints used for optimisation map to the notion of disparate impact is left as future work.

\noindent \textbf{Datasets:} The method is evaluated on \adult\ and \bank\ datasets.

\noindent \textbf{Fairness metrics:} Disparate impact, Disparate mistreatment.

\paragraph{\cite{grariFairAdversarialGradient2019}} proposed another adversarial method, \textit{Fair Adversarial Gradient Tree Boosting}, which again aims to make an accurate prediction, this time with Gradient Tree Boosting, whilst ensuring an adversarial cannot predict the sensitive attribute. This is done by integrating a fairness penalty into the loss function of the gradient tree boosting model.


\noindent \textbf{Benefit:} Results suggest that the method does not decrease performance significantly. For both fairness metrics, this method shows similar reduction in unwanted bias as Adversarial Learning \citep{zhangMitigatingUnwantedBiases2018}, Mechanisms For Fair Classification \citep{zafarFairnessConstraintsMechanisms2017} and Geometric Repair \citep{feldmanCertifyingRemovingDisparate2015,feldmanComputationalFairnessPreventing2015}, but improved accuracy over the three datasets.

\noindent \textbf{Limitations:}
\begin{enumerate}
    \item There is a risk of positive discrimination, i.e.  favouring individuals from the historically disadvantaged group over others. For example, females may be  chosen over men by the decision-maker, which may be fair statistically but not contextually, and in some cases, illegal \citep{wachterBiasPreservationMachine2021}.
    \item The method can only be applied to Gradient Tree Boosting models.
\end{enumerate}

\noindent \textbf{Datasets:} The method is evaluated on \adult, \bank, and \compas\ datasets.

\noindent \textbf{Fairness metrics:} Equalised odds, Demographic parity.

\paragraph{\cite{quadriantoRecyclingPrivilegedLearning2017}} proposed \textit{Privileged Learning \& Distribution Matching}. Privileged learning \citep{vapnikNewLearningParadigm2009} reduces the number of attributes needed to train an accurate classifier by using sensitive attributes to predict the slack variables, whilst distribution matching \citep{sharmanskaLearningMistakesOthers2016} minimises the distance between the distributions of two datasets.


\noindent \textbf{Benefits:}
\begin{enumerate}
    \item The method can be applied to datasets with multi-value and numerical sensitive attributes (e.g. age, income).
    \item In contrast to other methods \citep{fishConfidenceBasedApproachBalancing2016,lohiaBiasMitigationPostprocessing2019,kamiranDecisionTheoryDiscriminationAware2012}, the method does not use the sensitive attributes at test time, which may be realistic and an interesting consideration in some applications.
    \item The method allows the choice to optimise for equalised odds, equal opportunity, or demographic parity, similarly to \textit{Adversarial Learning} \citep{zhangMitigatingUnwantedBiases2018}.
\end{enumerate}

\noindent \textbf{Limitation:} Optimising with respect to one fairness metric can be at a detriment to a different metric. The results show that the method reduces equal opportunity but increases the equalised odds between the privileged and unprivileged groups.

\noindent \textbf{Datasets:} The method is evaluated on \adult\ and \compas\ datasets.

\noindent \textbf{Fairness metrics:} Equalised odds, Equal opportunity, Demographic parity.

\paragraph{\cite{onetoTakingAdvantageMultitask2019}} introduced \textit{Fair Multitask Learning} which splits the classification task into smaller tasks to learn group specific classifiers that leverage information between the privileged and unprivileged groups. If the sensitive attribute is sex, two classifiers would be learnt, one for male individuals and one for female individuals.


\noindent \textbf{Limitation:} Improving the fairness is at a detriment to the performance of a decision-making system. The results demonstrate that removing the sensitive attribute from training increases the fairness but decreases accuracy.

\noindent \textbf{Datasets:} The method is tested on \adult\ and \compas\ datasets.

\noindent \textbf{Fairness metrics:} Equalised odds, Equal opportunity.

\paragraph{\cite{iosifidisAdaFairCumulativeFairness2019}} proposed \textit{AdaFair} which is inspired by AdaBoost \citep{schapireBriefIntroductionBoosting1999} by using a cumulative approach to fairness and creating the best classifier from an ensemble of weighted trained classifiers based on the fairness and performance they achieve. Cumulative fairness, which is the sum of false positive and false negative rates over rounds of boosting, is shown to greatly outperform non-cumulative calculations of fairness.

\noindent \textbf{Limitations:}
\begin{enumerate}
    \item The distribution of the dataset used for the testing has an impact on the fairness value. Equalised odds is lower on datasets with higher original imbalance with respect to the sensitive attribute (e.g, \bank\ and \kdd), but yields worse performance on balanced datasets (e.g. \compas).
    \item Convergence of the method is not guaranteed and the number of iterations it may take to reach desired fairness cannot be calculated. These properties are left for future work.
\end{enumerate}

\noindent \textbf{Datasets:} The method is tested on \adult, \bank, \compas, and \kdd\ datasets.

\noindent \textbf{Fairness metrics:} Equalised odds.

\subsection{Post-processing Methods} \label{subsec:postprocess}

Post-processing bias mitigation methods involve swapping the classifications of individuals or adapting the output from the classifier (e.g. the posterior probabilities or labels of decision trees) to minimise metric values. Table \ref{tab:post-table} shows the collated post-processing methods that focus on removing unwanted bias from the classifications, along with the datasets and metrics used.


A limitation of post-processing methods is that they can be easily manipulated to ensure some existing fairness metric is satisfied, for example by swapping the classifications of random individuals to ensure an equal number of positive and negative classifications across sensitive groups.

\begin{table}[ht]
\begin{adjustbox}{width=1\textwidth}
    \begin{tabular}{lllll}
        \textbf{Method} & \textbf{Source}  & \textbf{Datasets} & \textbf{Fairness Metrics} & \textbf{Performance} \\
        & & & & \textbf{Metrics} \\
        \hline
        \hline
        Confidence Based Approach & \cite{fishConfidenceBasedApproachBalancing2016} & \adult , \german , \bank & Own fairness metric: & Accuracy \\
        & & & (RRB) & \\
        \hline
        Reject Option Based & \cite{kamiranDecisionTheoryDiscriminationAware2012} & \adult,  & Disparate impact & Accuracy \\
        Classification (ROC) & & \communities & & \\
        \hline
        Discrimination-Aware & \cite{kamiranDecisionTheoryDiscriminationAware2012} & \adult, & Disparate impact & Accuracy \\
        Ensemble (DAE) & & \communities & & \\
        \hline
        Individual and Group Debiasing & \cite{lohiaBiasMitigationPostprocessing2019} & \adult, \german, & Dwork IF, & Accuracy \\
        & & \compas & Disparate impact & \\
        \hline
        Relabelling Decision Trees & \cite{kamiranDiscriminationAwareDecision2010} & \adult, \dutch & Disparate impact & Accuracy \\
        \hline
        Modifying Naive Bayes & \cite{caldersThreeNaiveBayes2010} & \adult  & Disparate impact & Accuracy \\ \\
    \end{tabular}
    \end{adjustbox}
    \caption{Table of collated post-processing bias mitigation methods.}

    \label{tab:post-table}
\end{table}

\paragraph{\cite{fishConfidenceBasedApproachBalancing2016}} proposed a \textit{Confidence Based Approach} which aims to achieve fairness by moving the decision boundary between the positive and negative classifications. They define their own fairness metric, \textit{resilience to random bias (RRB)}, which is used to evaluate the method with respect to random bias introduced in the dataset, as opposed to the sensitive attribute. A new random attribute is added to the individuals in the dataset that does not correlate with the labels. The labels in the training dataset are then changed to include bias against a particular value of the new attribute, simulating unwanted bias in the system. Two models are trained, one with the adapted training dataset and one with the original dataset. RRB is quantified as the proportion of the classifications that are the same in the adapted model as in the original.


\noindent \textbf{Benefit:} The authors provide results with theoretical justifications, showing that post-processing methods are useful in mitigating unwanted bias, with minimal impact to the classifier's performance.

\noindent \textbf{Limitation:} The relation between the newly created fairness metric and existing metrics is not explained. The authors acknowledge that their metric represents a simple interpretation of bias and optimising for this measure is necessary but may not be sufficient to represent fairness overall.

\noindent \textbf{Datasets:} The method is evaluated on \adult, \german, and \bank\ datasets.

\noindent \textbf{Fairness metrics:} Own fairness metric (RRB).

\paragraph{\cite{kamiranDecisionTheoryDiscriminationAware2012}} proposed two decision theory based approaches for post-processing bias mitigation: \textit{Reject Option Based Classification (ROC)} and \textit{Discrimination-Aware Ensemble (DAE)}. \textit{ROC} exploits the probabilities of possible classifications of an individual to decide whether to change the label of that individual. In other words, depending on how confident the classifier is on its decision, the post-processing method will decide whether to change the decision in a way that reduces discrimination with respect to the sensitive attribute. For example, if the classifier gives a negative classification for an individual with the attribute value female with a confidence of 0.55, the classification will be changed to a positive one as that probability is within the critical region defined.

\textit{DAE} uses the differences between the classifications from a classifier ensemble to decide which individual's classifications
to change. If the classifications differ for an individual, the classification will be made positive if the individual is in the unprivileged group and made negative if the individual is in the privileged group.

\noindent \textbf{Benefit:} The results show that both methods decrease disparate impact without decreasing the performance significantly.

\noindent \textbf{Limitation:} ROC can only be applied to probabilistic classifiers and assumes access to the posterior probabilities of the classifications.

\noindent \textbf{Datasets:} The methods are tested on \adult\ and \communities\ datasets.

\noindent \textbf{Fairness metrics:} Disparate impact.

\paragraph{\cite{lohiaBiasMitigationPostprocessing2019}} proposed \textit{Individual and Group Debiasing} which extends ROC \citep{kamiranDecisionTheoryDiscriminationAware2012} but considers individual fairness as well as group fairness. The method identifies individuals with low individual fairness and swaps these individual's classifications based on the confidence of the classification.


\noindent \textbf{Limitation:} Similarly to other post-processing methods, satisfying individual fairness is more complex than group fairness and can be computationally costly to implement, especially as the number of individuals and sensitive attributes increases \cite{agarwalReductionsApproachFair2018}.

\noindent \textbf{Datasets:} The method is evaluated on \adult, \german\ and \compas\ datasets.

\noindent \textbf{Fairness metrics:} Dwork IF, Disparate impact.

\paragraph{\cite{kamiranDiscriminationAwareDecision2010}} proposed \textit{Relabeling Decision Trees} which swaps labels of leaves in the decision tree if doing so decreases disparate impact on the classifications, whilst maintaining or improving accuracy. This method is categorised as a post-processing method by its authors as it is applied after the model has been trained, however it involves changing the model and is specific to decision trees so is considered an intra-processing method by \cite{hortBiasMitigationMachine2022}.

\noindent \textbf{Benefit:} There is no detriment to the performance of the system when using this method. The results show a noticeable reduction in the disparate impact score without a loss in accuracy on the datasets evaluated.

\noindent \textbf{Limitation:} This method can only be applied to Decision Tree models.

\noindent \textbf{Datasets:} The method is evaluated on \adult\ and \dutch\ datasets.

\noindent \textbf{Fairness metrics:} Disparate impact.

\paragraph{\cite{caldersThreeNaiveBayes2010}} introduced a post-processing method for Naive Bayes classifiers.
\textit{Modifying Naive Bayes} is a model-specific \emph{post-processing} method that changes the probability of an individual in the unprivileged group with a positive class by modifying the Naive Bayes model's probabilities. It is considered an intra-processing method \cite{hortBiasMitigationMachine2022} as it post-processes the classifications but still requires the posterior probabilities thus it is included in Table~\ref{tab:post-table}.

\noindent \textbf{Limitation:} The method can only be applied to Naive Bayes models.

\noindent \textbf{Datasets:} The method is evaluated on the \adult\ dataset.

\noindent \textbf{Fairness metrics:} Disparate impact.

\section{Recommendations}
\label{discussion}

In this section, we review the limitations highlighted in Section~\ref{methods} and make recommendations for the development of future bias mitigation methods for binary classification decision-making systems.

\subsection*{Generalisability}

Most works on fairness in binary classification involve creating new bias mitigation methods that improve the fairness of the system according to some existing metric, and evaluating them using an arbitrarily chosen model and dataset. These methods usually focus on a single identified sensitive attribute (as the the majority of existing fairness metrics do not allow for multiple sensitive attributes) and on this attribute being binary as opposed to multi-value or numerical (e.g. only allows for the attribute of race having values ``white'' or ``non-white''). However, there can be multiple sensitive attributes (e.g. race) and trying to remove the unwanted bias with respect to one attribute may increase the unwanted bias towards another attribute~\citep{yuFairBalanceImprovingMachine2021}.

The specificity of these methods means they \textbf{do not generalise, making it hard to assess in what applications they would be useful.} Furthermore, there is currently no consensus of whether to choose a pre-, in-, or post-processing method to mitigate a decision-making system. The key factor for this decision is whether there is access to the model, in which case an in-processing method should be chosen rather than a pre- or post-processing method.

The main disadvantage of pre-processing methods is that they can be intrusive and involve editing the training data. Within the in-processing category, there are model-agnostic and model-specific methods, however, the disadvantage of these methods is that they require access to the model, which is not always possible. The disadvantage of post-processing methods is that they can easily manipulate the calculated fairness metrics without improving fairness. For example, it is possible to fully optimise for a fairness metric by simply swapping the classifications of some individuals in the unprivileged group to positive to have an equal number of positive and negative classifications for the privileged and unprivileged groups.

These findings contribute to the recommendation that \textbf{work on bias mitigation methods should also specify scenarios in which the methods may be applicable}.

\subsection*{Justification of Metrics}

Currently, fairness metrics are used to justify the improved fairness of a system, similarly to how performance metrics are used. Reducing fairness to an optimisation problem risks removing the key contextual importance of fairness. The methods created, often algorithmic, do not mention potential use cases or criteria for potential uses \citep{weinbergRethinkingFairnessInterdisciplinary2022}. This can make it difficult to evaluate when or where to use the different methods and whether real-life decision-making systems meet the technical criteria of a bias mitigation method (e.g. binary classification, access to the model).

Improving fairness can greatly impact performance. Due to the variety of fairness metrics, some conflicting with others, the theoretical relation between fairness and performance ought to be investigated more thoroughly. Methods that include parameters to decide the desired performance-fairness trade-off are useful in providing a bias mitigation method that may be applicable to multiple scenarios.

Thus, we recommend that \textbf{justifications are provided for the use of specific metrics} and how these relate to the context and application investigated.

\subsection*{Explainability}

Instead of simply using existing metrics to detect unwanted bias or evaluate the impact of bias mitigation methods, we recommend developing methods that focus on transparency and explainability, by  \textbf{explaining the unwanted bias detected in decision-making systems}. In particular, the fact that the fairness of a system has improved by e.g. 20\% after applying some bias mitigation method means very little to the individuals impacted, thus explaining unwanted bias is considered more useful compared to providing a number representing the unwanted bias \citep{vannuenenTransparencyWhomAssessing2020}.

\subsection*{Access to sensitive data}

There is rarely a discussion around the ethics and legality of the bias mitigation methods created. For example, data protection, specifically the use of protected attributes for use in prediction algorithms, is not discussed and many methods assume access to the protected attributes. However, this is not applicable in real-life applications due to data protection and equality treatment laws.\footnote{\url{https://www.legislation.gov.uk/ukpga/2010/15/contents}} \cite{haeriCrucialRoleSensitive2020} recognise the crucial role of protected attributes in being able to identify unwanted bias but that processing them is not allowed under Article 9 of \citep{gdprGeneralDataProtection2016}.

\subsection*{Addressing positive discrimination}

Many methods inadvertently cause positive discrimination which is where individuals in the unprivileged group are favoured over those in the privileged group. Under UK law, positive discrimination (or ``affirmative action'') is illegal. The potential for positive discrimination is rarely discussed, however, reducing discrimination without enforcing positive discrimination \citep{mancuhanCombatingDiscriminationUsing2014} remains an important consideration. Thus, \textbf{considering the legality of new bias mitigation methods} should be a key aspect.

Overall, when deciding whether to apply a bias mitigation method to a decision-making system, the recommendations highlighted should be considered to ensure that the method has the right impact on the application and individuals.

\section{Conclusion}
\label{conclusion}

In this article, we surveyed bias mitigation methods for binary classification decision-making systems and compared their strengths and limitations. We split the methods into pre-, in-, and post-processing methods, and reported the datasets and metrics used in evaluation. Our survey differs from existing work by critically analysing existing bias mitigation methods and providing recommendations for the development of future bias mitigation methods for binary classification decision-making systems. The recommendations made include ensuring generalisability of methods, justifying the use of existing metrics for their evaluation, exploring transparency and explainability, and considering the legality of bias mitigation methods.

\section*{Acknowledgments}
This work was supported by the UK Research and Innovation Centre for Doctoral Training in Safe and Trusted Artificial Intelligence [grant number EP/S023356/1].\footnote{\url{www.safeandtrustedai.org}}

\vskip 0.2in
\bibliography{ms}

\end{document}

%% file: macros.tex
\newcommand{\adult}{Adult}

\newcommand{\dutch}{Dutch}

\newcommand{\communities}{Communities \& Crime}

\newcommand{\german}{German}

\newcommand{\ricci}{Ricci vs DeStefano}

\newcommand{\compas}{COMPAS}

\newcommand{\law}{Law}

\newcommand{\iris}{Iris}

\newcommand{\diabetes}{Diabetes}

\newcommand{\census}{Census}

\newcommand{\bank}{Bank}

\newcommand{\default}{Default}

\newcommand{\wiki}{Wiki}

\newcommand{\student}{Student}

\newcommand{\health}{Health}

\newcommand{\kdd}{KDD}

\newcommand{\define}[2]{\begin{flushleft} \textbf{#1} {#2} \end{flushleft}}